\documentclass[letterpaper, 10 pt, conference]{ieeeconf}  % Comment this line out
\IEEEoverridecommandlockouts                              % This command is only
\usepackage[english]{babel}
\usepackage[T1]{fontenc}

\usepackage[nolist]{acronym}
\usepackage[cmex10]{amsmath}
\usepackage{amssymb}
\usepackage{bm}
\usepackage{booktabs}
\usepackage{color}
\usepackage{float}
\usepackage{graphicx}
\usepackage{hyperref}
\usepackage{import}
\usepackage{mathtools}
\usepackage{multirow}
\usepackage{outline}
\usepackage{paralist}
\usepackage{siunitx}
\usepackage{stmaryrd}
\usepackage{systeme}
\usepackage{units}
\usepackage{url}
\usepackage{tikz}
\usepackage{booktabs}
\usetikzlibrary{tikzmark}
\usepackage{amsmath}
\usepackage{caption}
\usepackage{subcaption}
\usepackage{booktabs}
\usepackage{wrapfig}
\title{\LARGE \bf Zero-shot Policy Learning with Spatial Temporal Reward Decomposition on Contingency-aware Observation }
\author{Huazhe Xu$^1$$^*$, Boyuan Chen$^1$$^*$, Yang Gao$^2$, Trevor Darrell$^1$ \thanks{$^{1}$ UC Berkeley}
\thanks{$^{2}$ Tsinghua University}\thanks{$^{*}$ Equal Contribution}}

\begin{document}
\maketitle

\begin{abstract}
%Learning one policy that can generalize to unseen compositional tasks without further data collection and finetuning is a long-standing challenge for creating true intelligent agent. In this paper, we propose a model-based method that can obtain a policy that can generalize to new configurations without any interaction with the previously unseen environment. Given a set of low-quality exploration offline data, our 
%Learning one policy that can generalize to unseen compositional tasks without further data collection and finetuning is a long-standing challenge for creating true intelligent agent. In this paper, we propose a model-based method that can obtain a policy that can generalize to new configurations without any interaction with the previously unseen environment. Given a set of imperfect offline data, our method learns a neural function that can decompose the delayed sparse reward into particular regions in an egocentric observation as a per step reward. Based on such decomposed rewards, we further learn a dynamics model and use Model Predictive Control (MPC) to obtain a policy. Since the rewards are decomposed to finer-granularity observations, they are naturally generalizable to new environments that are composed of similar basic elements. On various tasks including grid-world, classical video game super mario bros and robotics continuous control, we demonstrate that the proposed method is generalizable to new configurations and background and outperform all the competitive baselines. Please refer to the project page for more visualized results.

It is a long-standing challenge to enable an intelligent agent to learn in one environment and generalize to an unseen environment without further data collection and finetuning. In this paper, we consider a zero shot generalization problem setup that complies with biological intelligent agents' learning and generalization processes. The agent is first presented with previous experiences in the training environment, along with task description in the form of trajectory-level sparse rewards. Later when it is placed in the new testing environment, it is asked to perform the task without any interaction with the testing environment. We find this setting natural for biological creatures and at the same time, challenging for previous methods. Behavior cloning, state-of-art RL along with other zero-shot learning methods perform poorly on this benchmark. Given a set of experiences in the training environment, our method learns a neural function that decomposes the sparse reward into particular regions in a contingency-aware observation as a per step reward. Based on such decomposed rewards, we further learn a dynamics model and use Model Predictive Control (MPC) to obtain a policy. Since the rewards are decomposed to finer-granularity observations, they are naturally generalizable to new environments that are composed of similar basic elements. We demonstrate our method on a wide range of environments, including a classic video game -- Super Mario Bros, as well as a robotic continuous control task. Please refer to the project page for more visualized results.\footnote{https://sites.google.com/view/sapnew/home}

\end{abstract}
\section{INTRODUCTION}
\label{sec:Introduction}
While deep Reinforcement Learning (RL) methods have shown impressive performance on video games~\cite{mnih2015human} and robotics tasks~\cite{schulman2015high, lillicrap2015continuous}, they solve each problem \textit{tabula rasa}. Hence, it will be hard for them to generalize to new tasks without re-training even due to small changes. However, humans can quickly adapt their skills to a new task that requires similar priors \textit{e.g.} physics, semantics and affordances to past experience. The priors can be learned from a spectrum of examples ranging from perfect demonstrative ones that accomplish certain tasks to aimless exploration.  

\begin{figure}[t]
    \centering
    \includegraphics[width=0.48\textwidth]{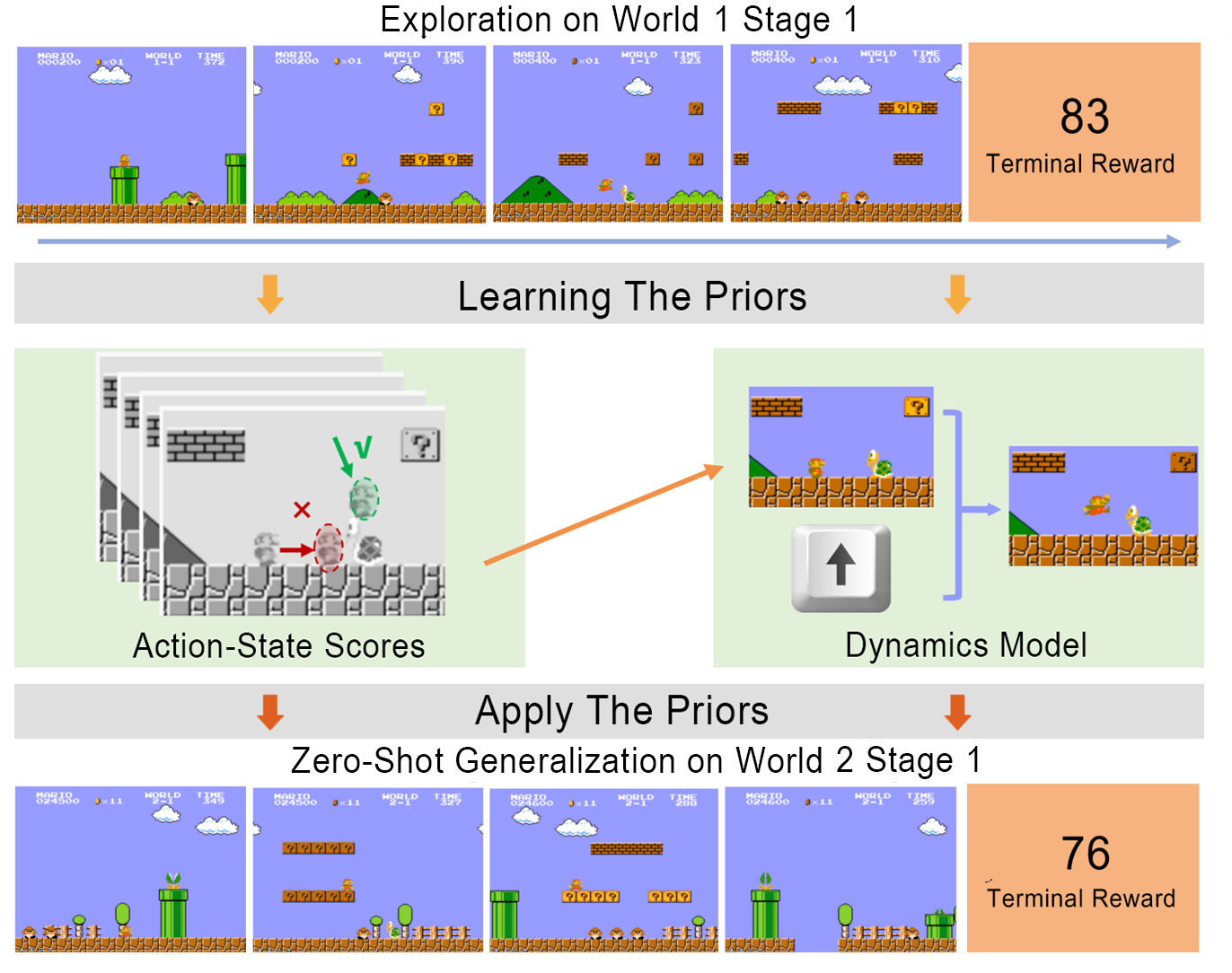}
    \caption{\small Illustrative figure: An agent is learning priors from exploration data from World 1 Stage 1 in Nintendo Super Mario Bros game. In this paper, the agent focuses on learning two types of priors: learning an action-state preference score for contingency-aware observation and a dynamics model. The action-state scores on the middle left learns that approaching the ``Koopa'' from the left is undesirable while from the top is desirable. On the middle right, a dynamics model can be learned to predict a future state based on the current state and action. The agent can apply the priors to a new task World 2 Stage 1 to achieve reasonable policy with zero shot.}
    \label{fig:concept}
    \vskip -0.5cm
\end{figure}

 A parameterized intelligent agent ``Mario'' who learns to reach the destination in the upper level in Figure~\ref{fig:concept} would fail to do the same in the lower new level because of the change of configurations and background, \textit{e.g.} different placement of blocks, new monsters. When an inexperienced human player is controlling the Mario to move it to the right in the upper level, it might take many trials for him/her to realize the falling to a pit and approaching the ``koopa''(turtle) from the left are harmful while standing on the top of the ``koopa''(turtle) is not. However, once learned, one can infer similar mechanisms in the lower level in Figure~\ref{fig:concept} without additional trials because human have a variety of priors including the concept of object, similarity, semantics, affordance, etc~\cite{gibson2014ecological, dubey2018investigating}. In this paper, we teach machine agents to realize and utilize useful priors  from exploration data in the form of decomposed rewards to generalize to new tasks without fine-tuning.

To achieve such zero-shot generalizable policy, a learning agent should have the ability to understand finer-granularity of the observation space e.g. to understand the value of a ``koopa'' in various configuration and background. However, these quintessentially human abilities are particularly hard for learning agents because of the lack of temporally and spatially fine-grained supervision signal and contemporary deep learning architectures are not designed for compositional properties of scenes. 
Many recent works rely heavily on the generalization ability of neural networks learning algorithms without capturing the compositional nature of scenes. 

In this work, we propose a method, which leverages imperfect exploration data that only have terminal sparse rewards to learn decomposed rewards on specific regions of an observation and further enable zero-shot generalization with a model predictive control (MPC) method. 
Specifically, given a batch of trajectories with terminal sparse rewards, we use a neural network to assign a reward for the contingency-aware observation $o_t$ at timestep $t$ so that the aggregation of the reward from each contingency-aware observation $o_t$ can be an equivalence of the original sparse reward. We adopt the contingency-aware observations~\cite{choi2018contingency} that enables an agent to be aware of its own location. Further, we divide the contingency-aware observation into $K$ sub-regions to obtain more compositional information. To further enable actionable agents to utilize the decomposed score, a neural dynamics model can be learned using self-supervision. We show that how an agent can take advantage of the decomposed rewards and the learned dynamics model with planning algorithms~\cite{mayne2000constrained}. Our method is called SAP where ``S'' refers to scoring networked used to decompose rewards,``A'' refers to the aggregation of per step rewards for fitting the terminal rewards, and ``P'' refers to the planning part.  

% Readers may argue that learning a dense score for every interaction step is reminiscent of Inverse Reinforcement Learning~\cite{ng2000algorithms,abbeel2004apprenticeship}. The distinctions between the proposed method and IRL are twofold: First, instead of learning a reward function of state $s$, we learn a scoring function of a egocentric state $s_l$ and an action $a$, which is sufficiently rich in a physical environment and experimentally can generalize well.  Second, with the scoring function in hand, we use a dynamics model learned from passive data to obtain the actual policy in a model-based manner while IRL needs to re-train an agent that can be as data inefficient as model-free RL. However, IRL can have difficulty learning a useful model because the expensive expert demonstrations usually only cover a small portion of the true dynamics. 
%Third, we eliminate the assumption of expensive expert demonstrations with the cost of adding a relatively economical sparse reward in the end. This elimination not only reduces the cost for data collection, but also includes more diverse data to train a robust model.

The proposed scoring function, beyond being a reward function for planning, can also be treated as an indicator of the existence of objects that affect the evaluation of a trajectory. We empirically evaluate the decomposed rewards for objects extracted in the context of human priors and hence find the potential of using our method as an unsupervised method for object discovery.

In this paper, we have two major contributions. First, we demonstrate the importance of decomposing sparse rewards into temporally and spatially smaller observation for obtaining zero-shot generalizable policy. Second, we develop a novel instance that uses our learning-based decomposition function and neural dynamics model that have strong performance on various challenging tasks. 
%% INTRODUCTION
\section{Related Work}
\textbf{Zero-Shot Generalization}
\textbf{and Priors}
To generalize in a new environment in a zero-shot manner, the agent needs to learn priors from its previous experiences, including priors on physics, semantics and affordances. Recently, researchers have shown the importance of priors in playing video games~\cite{dubey2018investigating}. More works have also been done to utilize visual priors in many other domains such as robotics for generalization, etc.~\cite{wang2019deep, jang2018grasp2vec, devin2018deep, zhu2018object, du2019task}. \cite{keramati2018strategic, DBLP:journals/corr/abs-1903-01385, higgins2017darla} explicitly extended RL to handle object level learning. While our method does not explicitly model objects, we have shown that meaningful scores are learned for objects enabling SAP to generalize to new tasks in zero-shot manner. Recent works~\cite{sohn2018hierarchical, oh2017zero} try to learn compositional skills for zero-shot transfer, which is complementary to the proposed method.  

\textbf{Inverse Reinforcement Learning.} The seminal work~\cite{ng2000algorithms} proposed inverse reinforcement learning (IRL). IRL aims to learn a reward function from a set of expert demonstrations. IRL and SAP fundamentally study different problem --- IRL learns a reward function from \textit{expert} demonstrations, while our method learns from exploratory data that is not necessarily related to any tasks. There are some works dealing with violation of the assumptions of IRL, such as inaccurate perception of the state~\cite{bogert2015multi,wang2002latent, bogert2016expectation, choi2011inverse}, or incomplete dynamics model~\cite{syed2008game, bogert2015multi, levine2014learning, bagnell2007boosting, ng2004feature}; however, IRL does not study the case when the dynamics model is purely learned and the demonstrations are suboptimal. Recent work~\cite{xie2019vrgrasp} proposed to leverage failed demonstrations with model-free IRL to perform grasping tasks; though sharing some intuition, our work is different because of the model-based nature.  

\textbf{RL with Sparse Reward}
When only sparse rewards are provided, an RL agent suffers a harder exploration problem. Previous work~\cite{ng1999policy} studied the problem of reward shaping, \textit{i.e.} how to change the form of the reward without affecting the optimal policy. The scoring-aggregating part can be thought as a novel form of learning-based reward shaping. A corpus of literature~\cite{marthi2007automatic, grzes2010online, marashi2012automatic} try to learn the reward shaping automatically. However, the methods do not apply to the high-dimensional input such as image.
%, while our SAP framework could. 
One recent work RUDDER~\cite{arjona2018rudder} utilizes an LSTM to decompose rewards into per-step rewards. This method can be thought of an scoring function of the full state in our framework.

There are more categories of methods to deal with this problem: 
(1) Unsupervised exploration strategies, such as curiosity-driven exploration~\cite{pathak2017curiosity, schmidhuber1991possibility}, or count-based exploration~\cite{tang2017exploration, strehl2008analysis} (2) In goal-conditioned tasks one can use Hindsight Experience Replay~\cite{andrychowicz2017hindsight} to learn from experiences with different goals. (3) defining auxiliary tasks to learn a meaningful intermediate representations~\cite{dosovitskiy2016learning}. In contrast to previous methods, we %approach this problem by learning a scoring function for each timestep, based on the single terminal reward. This
effectively convert the single terminal reward to a set of rich intermediate representations, on top of which we can apply planning algorithms.
%The line of work uses a variety of methods to learn an accurate dynamics model ranging from Gaussian Process~\cite{ko2009gp}, time-varying linear models~\cite{levine2013guided, lioutikov2014sample, xie2016model}, mixture of gaussian models~\cite{khansari2011learning} to neural networks~\cite{hunt1992neural, tangkaratt2014model, kurutach2018model, chua2018deep, luo2018algorithmic, buckman2018sample}.  This paradigm has been applied to high dimensional space, such as simulated and real robotic applications~\cite{watter2015embed, finn2016deep, hafner2018learning}, and Atari games~\cite{kaiser2019model, weber2017imagination}.
Although model-based RL has been extensively studied, none of the previous work has explored the use of reward decomposition for zero-shot transfer.

% \textbf{Model-Based RL.}
% In the planning part of our SAP framework, we train a dynamics model. \textit{i.e.} under the umbrella of model-based algorithms~\cite{sutton1991dyna}. This idea has been widely studied in the area of robotics~\cite{deisenroth2013survey, deisenroth2011pilco, morimoto2003minimax, deisenroth2011learning}.  The line of work uses a variety of methods to learn an accurate dynamics model \cite{ko2009gp, levine2013guided, lioutikov2014sample, xie2016model, khansari2011learning,hunt1992neural, tangkaratt2014model, kurutach2018model, chua2018deep, luo2018algorithmic, buckman2018sample}.  This paradigm has been applied to high dimensional space, such as simulated and real robotic applications~\cite{watter2015embed, finn2016deep, hafner2018learning}, and Atari games~\cite{kaiser2019model, weber2017imagination}.We combine model-based approach with learning the dense scores from sparse signals.
\section{Problem Statement and Method}
\subsection{Problem Formulation}
To be able to generalize better in an unseen environment, an intelligent agent should understand the consequence of its behavior both spatially and temporally. In the RL terminology, we propose to learn rewards that correspond to observations spatially and temporally from its past experiences. To facilitate such goals, we formulate the problem as follows.

Two environments $\mathcal{E}_1, \mathcal{E}_2$ are sampled from the same task distribution.  $\mathcal{E}_1$ and $\mathcal{E}_2$ share the same physics and goals but different configurations. e.g. different placement of objects, different terrains.

The agent is first presented with a bank of exploratory trajectories $\left\{\mathbf{\tau}_i\right\}, i=1, 2 \cdots N$ collected in training environment $\mathcal{E}_1=(\mathcal{S}, \mathcal{A}, p)$. Each $\mathbf{\tau}_i$ is a trajectory and $\mathbf{\tau}_i = \{(\mathbf{s}_t, \mathbf{a}_t)\}, t=1, 2 \cdots K_i$. These trajectories are random explorations in the environment. Note that the agent only learns from the bank of trajectories without further environment interactions, which mimics human utilizing only prior experiences to perform a new task. We provide a scalar terminal evaluation $r(\tau)$ of the entire trajectory when a task $\mathcal{T}$ is specified.

At test time, we evaluate task $\mathcal{T}$ using zero extra interaction in the new environment,  $\mathcal{E}_2=(\mathcal{S}', \mathcal{A}, p')$. We assume identical action space. There is no reward provided at test time. In this paper, we focus on locomotion tasks with object interactions, such as Super Mario running with other objects in presence, or a Reacher robot acting with obstacles around.

% With the task specification, we ask the agent to finish the new task $\mathcal{T'}$ in the test environment $\mathcal{E}_2$ without additional interaction. The performance is evaluated by the in-environment task final evaluation score $r(\mathbf{\tau})$, which is the same as how we specify the new task. 
\subsection{Spatial Temporal Reward Decomposition} 
%\textcolor{red}{ Currently I think the method section might lack some details, probably a new reader might find some details missing, let's fill it in. Especially considering the length of this section, it is a little bit less. }
% We propose the TBD framework with two stages. In the scoring stage, we propose to learn a per step neural scoring function $\mathrm{S_\theta}$ that scores a egocentric region, a subset of observation space that surrounds the agent. The egocentric region scores are aggregated along the trajectory to fit the terminal sparse reward. With the neural scoring function trained, we will get a decomposed per step reward with new observations. In the policy stage, this scoring function can be combined with planning and reinforcement learning algorithms to train a policy. In the zero-shot setting, we learn a dynamics model $\mathcal{M}_\phi$ to approximate the true transition $\mathrm{p}(\cdot |\mathbf{s}, \mathbf{a})$ based on past experience in $\mathcal{E}_1$. We then use Model Predictive Control algorithm as the final policy on the new environment $\mathcal{E}_2$.  

\begin{figure*}[h!]
\centering
\includegraphics[width=1.0\textwidth]{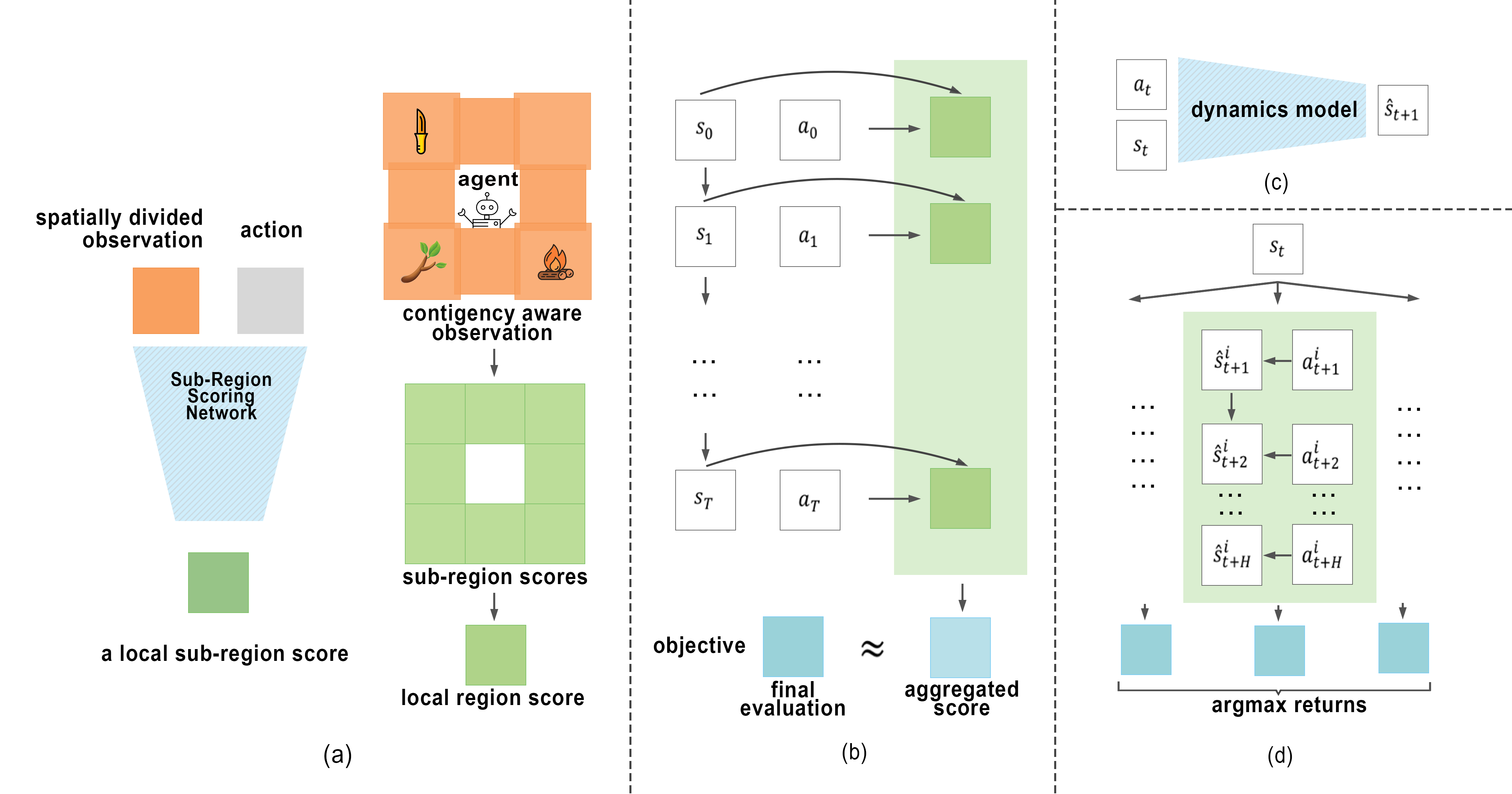}
  \caption{An overview of the SAP method. (a) For each time step, a scoring network scores contigent sub-regions conditioned on action. (b) we aggregate the prediction over all time steps to fit terminal reward (c)\&(d) describe the dynamics learning and planning in mpc in the policy stage.}
  \label{fig:method}
  \vskip -0.4cm
\end{figure*}

In this section, we introduce the method to decompose the terminal sparse reward into specific time step and spatial location. First, we introduce the temporal decomposition and then discuss the spatial decomposition.

\textbf{Temporal Reward Decomposition}
The temporal reward decomposition can be described as  $\mathrm{S_\theta}(\mathrm{W}(\mathbf{s}_t), \mathbf{a}_t)$. Here $\theta$ denotes parameters in a neural network,  $\mathrm{W}$ is a function that extracts contingency-aware observations from states.  Here, contingency-aware means a subset of spatial global observation around the agent, such as pixels surrounding a game character or voxels around end-effector of a manipulator. We note that the neural network's parameters are shared for every contingency-aware observation. Intuitively, this function measures how well an action $\mathbf{a}_t$ performs on this particular state, and we refer to $S_\theta$ as the \textit{scoring function}.

To train this network, we aggregate the decomposed rewards $ \mathrm{S}_\theta(\mathrm{W}(\mathbf{s}_t), \mathbf{a}_t)$ for each step into a single aggregated reward $\mathnormal{J}$, by an aggregating function $\mathrm{G}$:
$$\mathnormal{J}_\theta(\mathbf{\tau}) = \mathrm{G}_{(\mathbf{s}_t, \mathbf{a}_t) \in \mathbf{\tau}}( S_\theta(W(s_t), a_t))$$
The aggregated reward $\mathnormal{J}$ are then fitted to the sparse terminal reward. In practice, $\mathrm{G}$ is chosen based on the form of the sparse terminal reward, \textit{e.g.} a max or a sum function. In the learning process, the $\mathrm{S}_\theta$ function is learned by back-propagating errors between the terminal sparse reward and the predicted $\mathnormal{J}$ through the aggregation function. In this paper, we use $\ell_2$ loss that is: $$\min_{\theta} \frac{1}{2}(\mathnormal{J}_\theta(\mathbf{\tau})-r(\tau))^2$$

%In our method, we presume contingency-aware observation can be extracted for decomposition because in a physical environment, an agent could only interact with the world within its sensing capabilities. 

\textbf{Spatial Reward Decomposition}
An environment usually contains multiple objects. Those objects might re-appear at various spatial locations over time. To further assist the learned knowledge to be transferrable to the new environment, we take advantage of the compositionality of the environment by also decomposing the reward function spatially. More specifically, we divide the contingency-aware observation into smaller sub-regions. For example, in a Cartesian coordinate system, we can divide each coordinate independently and uniformly. With the sub-regions, we re-parametrize the scoring function as $\sum_{l \in \mathcal{L}} \mathrm{S}_\theta(\mathrm{W}_l(\mathbf{s}_t), \mathbf{a}_t)$, where $l$ is the index of the sub-regions and we retarget $S_\theta$ for the sub-region instead of the whole contingency-aware observation. The intuition is that the smaller sub-regions contains objects or other unit elements that are also building blocks of unseen environments. Such sub-regions become crucial later to generalize to the new environment. 

\subsection{Policy Stage}
%With a learned scoring function that decomposes sparse terminal reward to dense reward per step, we can learn a policy from the demonstration as in standard MDP setting. 
To solve the novel task with zero interaction, we propose to use planning algorithms to find optimal actions based on the learned scoring function and a learned dynamics model. As shown in the part (c) of Figure~\ref{fig:method}, we learn a forward dynamics model $\mathcal{M}_\phi$ based on the exploratory data with a supervised loss function. Specifically, we train a neural network that takes in the action $\mathrm{a}_t$, state $\mathrm{s}_t$ and output $\hat{\mathrm{s}}_{t+1}$, which is an estimate of $\mathrm{s}_{t+1}$. We use an $\ell_2$ loss as the objective:
$$\min_\phi \frac{1}{2}(\mathcal{M}_\phi(\mathrm{s}_t,\mathrm{a}_t)-\mathrm{s}_{t+1})^2$$

With the learned dynamics model and the scoring function, we solve an optimization problem using the Model Predictive Control (MPC) algorithm to find the best trajectory for a task $\mathcal{T}$ in environment $\mathcal{E}_2$. The objective of the optimization problem is to minimize $\mathnormal -\mathnormal{J}_\theta(\mathbf{\tau}')$. Here we randomly sample multiple action sequences up to length H, unroll the states based on the current state and the learned dynamics model while computing the cost with the scoring function. We select the action sequence with the minimal cost, and execute the first action in the selected sequence in the environment.

\subsection{Discussion on zero-shot generalization}
Although neural networks have some generalization capabilities, it is still easy to overfit to the training domain. Previous works~\cite{packer2018assessing} notice that neural network ``memorizes'' the optimal action throughout the training process. One can not expect an agent that only remembers optimal actions to generalize well when placed in a new environment. Our method does not suffer from the memorization issue as much as the neural network policy because it can come up with novel solutions, that are not necessarily close to the training examples, since our method learns generalizable model and scores that makes planning possible. The generalization power comes mainly from the smaller building blocks shared across environments as well as the universal dynamics model. This avoids the SAP method to replay the action of the nearest neighbour state in the training data. 

%The intuition of why our proposed SAP method works in the zero shot generalization case is that instead of running a trained neural network policy in the new environment, we run a planning algorithm on the fly.

%Although neural networks have some generalization capabilities, it is still easy to overfit to the training domain. Some previous works~\cite{todo} notice that neural network ``memorize'' the optimal action throughout the training process. One can not expect an agent that only remembers optimal actions to generalize well when placed in a novel environment. On the other hand, our method searches for a good action based on the learned scores on the fly. It is not limited to ``replaying'' the optimal action, but it can search for an appropriate one based on the specific situation of the testing environment. % SAP has a better zero-shot generalization ability because (1) it considers \textit{egocentric} states which can be shared among very different environments; (2) it only requires a good scoring function to have the correct order of scores rather than correct exact value.

Section \ref{sec:comparisons}, Figure~\ref{fig:mario train}~\ref{fig:mario test} compares our method and a neural network RL policy. It shows that in the training environment, RL policy is only slightly worse than our method, however, the RL policy performs much worse than ours in the zero shot generalization case. 

\section{Experiment}
% \textcolor{red}{I leave a lot of details in the robot experiment but omit a lot of details in the super mario experiments. So now it might seems a little bit unbalanced between the two set of experimtns. But I want to balance the two. Please add back some details in the super mario one and delete some from the robotics one, such that the level of details are not super differnet from each other. }

In this section, we study how well the proposed method performs compare to baselines, and the roles of the proposed temporal and spatial reward decompositions. We conduct experiments on two domains: a famous video game ``Super Mario Bros''~\cite{gym-super-mario-bros} and a robotics blocked reacher task~\cite{huang2019mapping}.

\subsection{Experiment on Super Mario Bros}
To evaluate our proposed algorithm in a challenging environment, we run our method and baseline methods in the Super Mario Bros environment. This environment features high-dimensional visual observations, which is challenging since we have a large hypothesis space. The original game has 240 $\times$ 256 image input and discrete action space with 5 choices. We wrap the environment following \cite{mnih2015human}. Finally, we obtain a $84 \times 84$ size 4-frame gray-scale stacked observation. The goal for an agent is to survive and go toward the right as far as possible. We don't have access to the dense reward for each step but the environment returns how far the agent moves towards target at the end of a trajectory as the delayed terminal sparse reward.

\subsubsection{Baselines}
\label{sec:comparisons}
We compare our method with various types of baselines, including state-of-art zero-shot RL algorithms, generic policy learning algorithms as well as oracles with much more environment interactions. More details can be found in the appendix. 

\texttt{Exploration Data} Exploration Data is the data from which we learn the scores, dynamics model and imitate. The data is collected from noisy sub-optimal version of policy trained using~\cite{pathak2017curiosity}. The average reward on this dataset is a baseline for all other methods. 

\texttt{Behavioral Cloning~\cite{pomerleau1989alvinn, bain1995framework}} Behavioral Cloning (BC) learns a mapping from a state to an action on the exploration data using supervised learning. We use cross-entropy loss for predicting the actions. 

\texttt{Model Based with Human Prior} Model Based with Human Priors method (MBHP) incorporates model predictive control with predefined human priors, which is +1 score if the agent tries to move or jump toward the right and 0 otherwise. 

\texttt{DARLA~\cite{higgins2017darla}} DARLA relies on learning a latent state representation that can be transferred from the training environments to the testing environment. It achieves this goal by obtaining a disentangled representation of the environment’s generative factors before learning to act. We use the latent representation as the observations for a behavioral cloning agent.  

\texttt{RUDDER~\cite{arjona2018rudder}} RUDDER proposes to use LSTM to decompose the delayed sparse reward to dense rewards. It first trains a function $f(\tau_{1:T})$ predict the terminal reward of a trajectory $\tau_{1:T}$. It then use $f(\tau_{1:t})-f(\tau_{1:t-1})$ as dense reward at step $t$ to train RL policies. We change policy training to MPC so this reward can be used in zero-shot setting. 

\texttt{Behavior Clone with Privilege Data} Instead of using exploratory trajectories from the environment, we collect a set of near optimal trajectories in the training environment and train a behavior clone agent from it. Note that this is not a fair comparison with other methods, since this method uses better performing training data. 

\texttt{RL curiosity} We use a PPO~\cite{schulman2017proximal} agent that is trained with curiosity driven reward~\cite{burda2018exploration} and the final sparse reward in the training environment. This also violates the setting we have as it interacts with the environment. We conduct this experiment to test the generalization ability of an RL agent.

\begin{figure*}[t]
    \centering
    \begin{subfigure}[b]{0.34\textwidth}
        \centering
        \includegraphics[width=\textwidth]{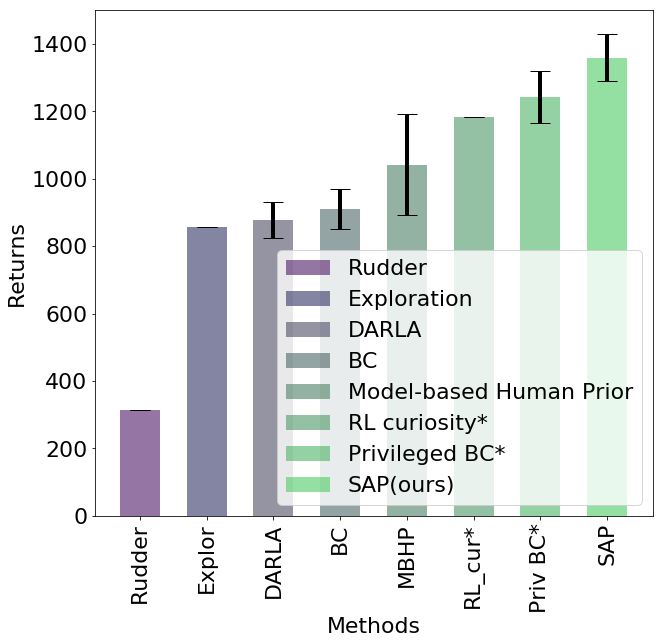}
        \caption{Mario train world returns}
        \label{fig:mario train} 
    \end{subfigure}%
    \begin{subfigure}[b]{0.34\textwidth}
        \centering
        \includegraphics[width=\textwidth]{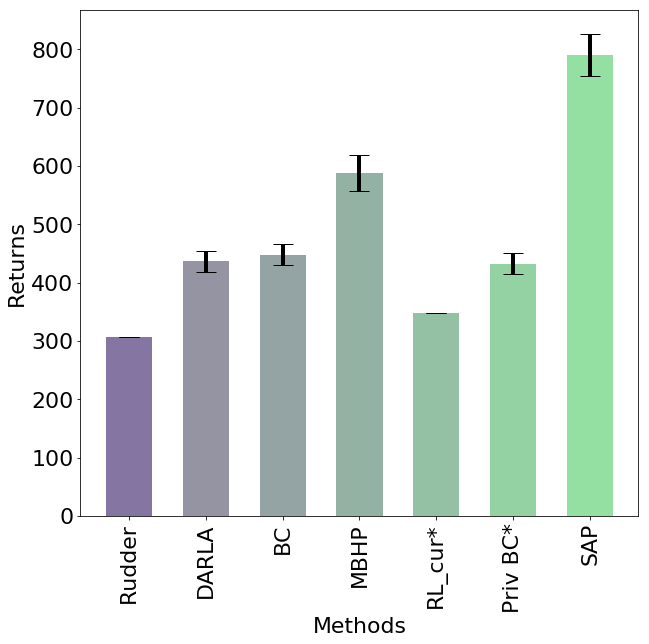}
        \caption{Mario test world returns}
        \label{fig:mario test}
    \end{subfigure}
    \begin{subfigure}[b]{0.3\textwidth}
        \centering
        \includegraphics[width=\textwidth]{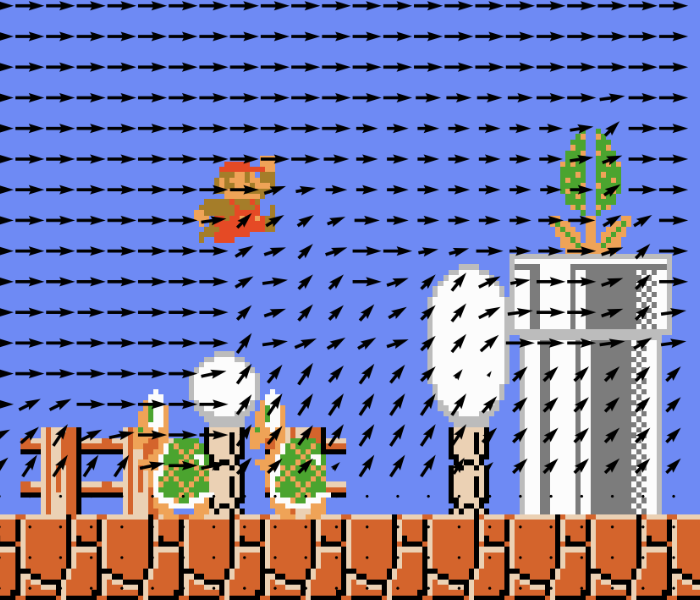}
        \vskip 0.75cm
        \caption{Visualized greedy actions from the learned scores in a new level(W5S1)}
        \label{fig:mario visualization}
    \end{subfigure}
    \caption{(a) \& (b) The total returns of different methods on the Super Mario Bros. %All the experiments are run with 50000 game steps on multiple episodes. 
    Methods with $*$ have privilege access to optimal data instead of sub-optimal data.
    Error bars are shown as 95\% confidence interval. %except for the given interaction data. 
    See Section \ref{sec:comparisons} for details. \label{fig:full} %Best view in color.
    }
    \label{fig:mario benchmark}
    \vskip -0.5cm
\end{figure*}

\subsubsection{Analysis}
Fig.~\ref{fig:mario benchmark}~\subref{fig:mario train} and Fig.~\ref{fig:mario benchmark}~\subref{fig:mario test} show how the above methods performs in the Super Mario Bros. The \texttt{Explore} baseline shows the average training trajectory performance. We found that \texttt{RUDDER} fails to match the demonstration performance. This can attributed to the LSTM in RUDDER not having sufficient supervision signals from the long horizon (2k steps) delayed rewards in the Mario environment. \texttt{DARLA} slightly outperforms the demonstration data in training. Its performance is limited by the unsupervised visual disentangled learning step, which is a hard problem in the complex image domain. \texttt{Behavior Cloning} is slightly better than the exploration data, but much worse than our method. \texttt{MBHP} is a model-based approach where human defines the per step reward function. However, it is prohibitive to obtain detailed manual rewards every step. In this task, it is inferior to SAP's learned priors. We further compare to two methods that unfairly utilize additional privileged information. The \texttt{Privileged BC} method trains on near optimal data, and performs quite well during training; however, it performs much worse during the zero shot testing. The similar trend happens with \texttt{RL Curiosity}, which has the privileged access to online interactions in the training environment. 

To conclude, we found generic model free algorithms, including both RL (\texttt{RL Curiosity}), and behavior cloning (\texttt{Privilege BC}) perform well on training data but suffer from severe generalization issue. The zero-shot algorithms (\texttt{DARLA}), BC from exploratory data (\texttt{BC}) and moded-based method (\texttt{MBHP}) suffer less from degraded generalization, but they all under-perform our proposed algorithm.

%\textbf{Analysis} In Figure.~\ref{fig:full}\subref{fig:main}, the results show that the proposed SAP model outperforms all the baselines with a large margin on the same level they are trained on. We believe that there are two major reasons for BC's unsatisfactory performance: 1. we only have access to the exploration data which is suboptimal for the task. 2. When seeing rare events in the game (the exploration data hardly reach the ending part), it fails to generalize. SAP also outperforms DARLA on both training and generalization tasks. We believe it is because observations can change dramatically between train and test levels. Thus, DARLA's learned disentangled representations can't generalize well enough in test level. Comparing the SAP model with the MBHP method, we demonstrate the learned priors can empower an agent with a better understanding of the world thus stronger performance. We believe rudder fails on the environment for two reasons: 1. Rudder is trained on 
%We show qualitative results in the subsection~\ref{sec:viz} to validate that the learned priors contain meaningful scores that leads to better actions. SAP also outperforms all the baselines in an unseen level without any finetuning (Fig.~\ref{fig:full}\subref{fig:generalization}), which proves that it can generalize well. \bibiY{There should be more convincing reasons that explains why our method is better than the DARLA and rudder baselines}

\subsubsection{Ablative Studies}
\label{sec:ablation}
\begin{table}
{\small
\centering
\caption{Ablation of the temporal and spatial reward decomposition. \label{table:visual_representation}}
\begin{tabular}{@{}lll@{}}\toprule
& {W1S1(train)} & {W2S1(test)} 
            \\ \midrule
SAP      & \textbf{1359.3} & \textbf{790.1}  \\
SAP w/o spatial     & 1258.0          & 737.0  \\
SAP w/o spatial temporal                 & 1041.3          & 587.9  \\
\bottomrule
\end{tabular}
\vskip -0.3cm
}
\end{table}

In order to have a more fine-grained understanding of the effect of our newly proposed temporal and spatial reward decomposition, we further conduct ablation studies on those two components. We run two other versions of our algorithm, removing the spatial reward decomposition and the temporal reward decomposition one at a time. More specifically, the \texttt{SAP w/o spatial} method does not divide the observation into sub-regions, but simply use a convolution network to approximate the scoring function. \texttt{SAP w/o spatial temporal} further removes the temporal reward learning part, and replace the scoring function with a human-designed prior. I.e. it is the same as the MBHP baseline. Please see appendix for more details. 

Table \ref{table:visual_representation} shows the result. We found that both the temporal and the spatial reward learning component contributes to the final superior performance. Without temporal reward decomposition, the agent either have to deal with sparse reward or use a manually specified rewards which might be tedious or impossible to collect. Without the spatial decomposition, the agent might not find correctly which specific region or object is important to the task and hence fail to generalize.  

\subsubsection{Visualization of learned scores}\label{sec:viz}
In this section, we qualitatively study the induced action by greedily maximizing one-step score. I.e. for any location on the image, we assumes the Super Mario agent were on that location and find the action $a$ that maximize the learned scoring function $S_\theta (W(s), a)$. We visualize the computed actions on World 5 Stage 1 (Fig.~\ref{fig:mario benchmark}~\subref{fig:mario visualization}) which is visually different from previous tasks. More visualization can be found in \footnote{https://sites.google.com/view/sapnew/home} In this testing case, we see that the actions are reasonable, such as avoiding obstacles and monsters by jumping over them, even in the face of previously unseen configurations and different backgrounds. However, the ``Piranha Plants'' are not recognized because all the prior scores are learned from W1S1 where it never appears. More visualization of action maps are available in our videos. Those qualitative studies further demonstrate that the SAP method can assign meaningful scores for different objects in an unsupervised manner. 
It also produces good actions even in a new environment.

\subsection{SAP on the 3-D robotics task}
\label{sec:3d}
In this section, we further study the SAP method to understand its property with a higher dimensional observation space. We conduct experiments in a 3-D robotics environment, BlockedReacher-v0. In this environment, a robot hand is initialized at the left side of a table and tries to reach the right. Between the robot hand and the goal, there are a few blocks standing as obstacles. The task is moving the robot hand to reach a point on $y=1.0$ as fast as possible. To test the generalization capability, we create four different configurations of the obstacles, as shown in Figure~\ref{fig:samples}. Figure~\ref{fig:samples} A is the environment where we collect exploration data from and Figure~\ref{fig:samples}~B, C, D are the testing environments. Note that the exploration data has varying quality, where many of the trajectories are blocked by the obstacles. We introduce more details about this experiment. 

\subsubsection{Environment.}\label{sec:robot_env_app}
In the Blocked Reach environment, we use a 7-DoF robotics arm to reach a specific point. For more details, we refer the readers to \cite{plappert2018multi}. We discretize the robot world into a $200 \times 200 \times 200$ voxel cube. For the action space, we discretize the actions into two choices for each dimension which are moving 0.5 or -0.5. Hence, in total there are 8 actions. We design four configurations for evaluating different methods as shown in Figure~\ref{fig:samples}. For each configurations, there are three objects are placed in the middle as obstacles. 

\subsubsection{Applying SAP on the 3D robotic task}
\label{sec:app:sap_robot}
We apply the SAP framework as follows. The original observation is a 25-dimensional continuous state and the action space is a 3-dimensional continuous control. They are discretized into voxels and 8 discrete actions as described in Appendix~\ref{sec:robot_env_app}. The scoring function is a fully-connected neural network that takes in a flattened voxel sub-region. We also train a 3D convolutional neural net as the dynamics model. The dynamics model takes in the contingency-aware observation as well as an action, and outputs the next robot hand location. With the learned scores and the dynamics model, we plan using the MPC method with a horizon of 8 steps.  

\subsubsection{Architectures for score function and dynamics model.}
For the score function, we train a 1 hidden layer fully-connected neural networks with 128 units. We use ReLu functions as activation except for the last layer. Note that the input 5 by 5 by 5 voxels are flattened before put into the scoring neural network.

For the dynamics model, we train a 3-D convolution neural network that takes in the contingency-aware observation (voxels),  action and last three position changes. The voxels contingent to end effector are encoded using three 3d convolution with kernel size 3 and stride 2. Channels of these 3d conv layers are 16, 32, 64, respectively. A 64-unit FC layer is connected to the flattened features after convolution. The action is encoded with one-hot vector connected to a 64-unit FC layer. The last three $\delta$ positions are also encoded with a 64-unit FC layer. The three encoded features are concatenated and go through a 128-unit hidden FC layer and output predicted change in position. All intermediate layers use ReLu as activation.   

% \subsubsection{Hyperparameters.}
% During training, we use an adam optimizer with learning rate 3e-4, $\beta_1=0.9$, $\beta_2 = 0.999$. The batchsize is 128 for score function training and 64 for dynamics model.  We use $horizon=8$ as our planning horizon. We use 2e-5 as the weight for matrices entry regularization.

\begin{table}[h]
    \centering
    \caption{Evaluation of SAP, MBHP and Rudder on the 3D Reacher environment. Numbers are 
    the avg. steps to reach the goal. The lower the better. Numbers in the brackets are the 95\% confidence interval. ``L'' denotes the learned dynamics, and ``P'' denotes the perfect dynamics.}
    \label{tab:robot}
    \begin{tabular}{l@{\hspace{0.32cm}}c@{\hspace{0.64cm}}c@{\hspace{0.64cm}}c@{\hspace{0.64cm}}l}
    \toprule
    &Config A&Config B&Config C& Config D\\ \hline
    SAP(L)&\textbf{97.53}[2.5]&\textbf{86.53}[1.9]&\textbf{113.3}[3.0]&\textbf{109.2}[3.0]\\
    MBHP(L)&124.2[4.5]&102.0[3.0]&160.7[7.9]&155.4[7.2]\\
    Rudder(L)&1852[198]&1901[132]&1933[148]&2000[0.0]\\
    \hline
    SAP(P)&\textbf{97.60}[2.6]&\textbf{85.38}[1.8]&\textbf{112.2}[3.1]&\textbf{114.1}[3.2]\\
    MBHP(P)&125.3[4.3]&102.4[3.0]&153.4[5.3]&144.9[4.8]\\
    Rudder(P)&213.6[4.2]&194.2[7.0]&208.2[5.6]&201.5[6.3]\\
    \bottomrule
    \end{tabular}
    \label{tb:robot benchmark}
    % \vskip -0.3cm
\end{table}

% \begin{table}[h]
%         \centering
%         \caption{Evaluation of SAP, MBHP and Rudder on the 3D Reacher environment. Numbers are 
%         the avg. steps to reach the goal. The lower the better. Numbers in the brackets are the 95\% confidence interval. ``L'' denotes the learned dynamics, and ``P'' denotes the perfect dynamics.}
%         \label{tab:robot}
%         \begin{tabular}{l@{\hspace{0.1cm}}c@{\hspace{0.2cm}}c@{\hspace{0.2cm}}c@{\hspace{0.2cm}}l}
%         \toprule
%         &Config A&Config B&Config C& Config D\\ \hline
%         SAP(L)&\textbf{97.53}[2.5]&\textbf{86.53}[1.9]&\textbf{113.3}[3.0]&\textbf{109.2}[3.0]\\
%         MBHP(L)&124.2[4.5]&102.0[3.0]&160.7[7.9]&155.4[7.2]\\
%         Rudder(L)&1852[198]&1901[132]&1933[148]&2000[0.0]\\
%         \hline
%         SAP(P)&\textbf{97.60}[2.6]&\textbf{85.38}[1.8]&\textbf{112.2}[3.1]&\textbf{114.1}[3.2]\\
%         MBHP(P)&125.3[4.3]&102.4[3.0]&153.4[5.3]&144.9[4.8]\\
%         Rudder(P)&213.6[4.2]&194.2[7.0]&208.2[5.6]&201.5[6.3]\\
%         \bottomrule
%         \end{tabular}
%         \label{tb:robot benchmark}
%         % \vskip -0.3cm
% \end{table}

\begin{figure}
\centering
  \begin{subfigure}[b]{0.15\textwidth}
  \centering
  \includegraphics[width=\linewidth]{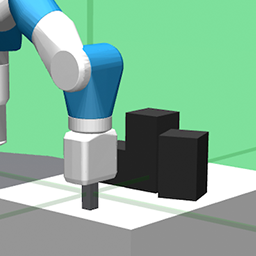}
  \caption{}\label{fig:robot0} 
\end{subfigure}
\begin{subfigure}[b]{0.15\textwidth}
  \centering
  \includegraphics[width=\linewidth]{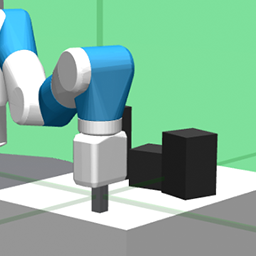}
  \caption{}\label{fig:robot1}
\end{subfigure}

\begin{subfigure}[b]{0.15\textwidth}
  \centering
  \includegraphics[width=\linewidth]{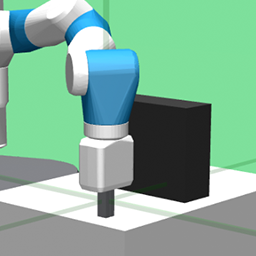}
  \caption{}\label{fig:robot2} 
\end{subfigure}
\begin{subfigure}[b]{0.15\textwidth}
  \centering
  \includegraphics[width=\linewidth]{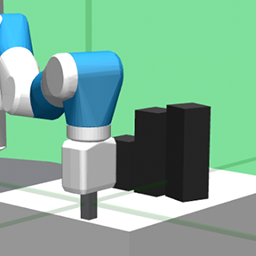}
  \caption{}\label{fig:robot3} 
\end{subfigure}
\caption{Four variants of the 3D robot reacher environments. See Section \ref{sec:3d} for details.}\label{fig:samples}
\vskip -0.4cm
\end{figure}

% \begin{wrapfigure}{R}{0.32\linewidth}
% \centering
% \includegraphics[width=\linewidth]{contents/env_sample/task_composed.png}
% \caption{Four variants of the 3D robot reacher environments. See Section \ref{sec:3d} for details.}\label{fig:samples}
% % \vskip -0.3cm
% \end{wrapfigure}

\subsubsection{Results}
We evaluate similar baselines as in the previous section that is detailed in ~\ref{sec:comparisons}. In Table~\ref{tb:robot benchmark}, we compare our method with the MBHP and RUDDER on the 3D robot reaching task. We found that our method needs significantly fewer steps than the two baselines, in both training environment and testing ones. We find that SAP significantly moves faster to the right because it learns a negative score for crashing into the obstacles. However, the MBHP method, which has +1 positive for each 1 meter moved to the right, would be stuck by the obstacles for a longer duration. When training with RUDDER, the arm also frequently waste time by getting stuck at obstacle. We found that our SAP model is relatively insensitive to the errors in the learned dynamics, such that the performance using the learned dynamics is close to that of perfect dynamics. These experiments show that our method can be applied to robotics environment that can be hard for some algorithms due to the 3-D nature. %Moreover, we demonstrate again that the SAP method perform much better than all other methods. %Moreover, we demonstrate again that using partial states (egocentric regions) with SAP generalize better than baselines. 

\section{Conclusion}
In this paper, we introduced a new method called SAP that aims to generalize in the new environment without any further interactions by learning the temporally and spatially decomposed rewards. We empirically demonstrate that the newly proposed algorithm outperform previous zero-shot RL method by a large margin, on two challenging environments, i.e. the Super Mario Bros and the 3D Robot Reach. The proposed algorithm along with a wide range of baselines provide a comprehensive understanding of the important aspect zero-shot RL problems.
\section{Acknowledgement}
The work has been supported by BAIR, BDD, and DARPA XAI. This work has been supported in part by the Zhongguancun Haihua Institute for Frontier Information Technology. Part of the work done while Yang Gao is at UC Berkeley. We also thank Olivia Watkins for discussion.

%% BIBLIOGRAPHY
{
\bibliographystyle{IEEEtran}
\bibliography{bibliography.bib}
}

\clearpage
\newpage
\appendix
\section{Experiment Specs}\label{appendix:specs}
\subsection{Hidden Reward Gridworld}
\subsubsection{Environment.}
% In Figure~\ref{fig:three graphs}\subref{fig:grid}, we visualize a sample of the gridworld environment.
In the gridworld environment, each entry correspond to a feature vector with noise based on the type of object in it. Each feature is a length 16 vector whose entries are uniformly sampled from $[0, 1]$. Upon each feature, we add a small random noise from a normal distribution with $\mu = 0, \sigma = 0.05$. The outer-most entries correspond to padding objects whose rewards are 0.  The action space includes move toward four directions up, down, left, right. If an agent attempts to take an action which leads to outside of our grid, it will be ignored be the environment.

\subsubsection{Architectures for score function and dynamics model.} 
We train a two layer fully connected neural networks with 32 and 16 hidden units respectively and a ReLU activation function to approximate the score for each grid. 

In this environment, we do not have a learned dynamics model.

\textbf{Hyperparameters.} During training, we use an adam optimizer with learning rate 1e-3, $\beta_1=0.9$, $\beta_2 = 0.999$. The learning rate is reduced to 1e-4 after 30000 iterations. The batchsize is 128. We use $horizon=4$ as our planning horizon.

\subsection{Super Mario Bros}\label{sec:mario_app}

\subsubsection{Environment.} We wrap the original Super Mario environments with additional wrappers. We wrap the action space into 5 discrete joypad actions, none, walk right, jump right, run right and hyper jump right. We follow~\cite{burda2018exploration} to add a sticky action wrapper that repeats the last action with a probability of 20\%. Besides this, we follow add the standard wrapper as in past work~\cite{mnih2015human}.

\subsubsection{Applying SAP on Super Mario Bros}
We apply our SAP framework as follows. We first divide the egocentric region around the agent into eight 12 by 12 pixel sub-regions based on relative position as illustrated in Figure.~\ref{fig:locations} in the Appendix. Each sub-region is scored by a CNN, which has a final FC layer to output a score matrix. The matrix has the shape dim(action) by dim(relative position), which are 5 and 8 respectively. Then an action selector and a sub-region selector jointly select row corresponding to the agent's action and the column corresponding to the relative position. The sum of all the sub-region scores forms the egocentric region score. Then we minimize the $\ell_2$ loss between the aggregated egocentric region scores along the trajectory and the terminal reward. In addition, we add an regularization term to the loss to achieve better stability for long horizon. We take the $\ell1$ norm of score matrices across trajectory to add to loss term. This encourages the scoring function to center predicted loss terms around 0 and enforce numerical sparsity. A dynamics model is also learned by training another CNN. The dynamics model takes in a 30 by 30 size crop around the agent, the agent's location as well a one-hot action vector.  Instead of outputting a full generated image, we only predict the future location of the agent recursively. We avoid video predictive models because it suffers the blurry effect when predicting long term future~\cite{lotter2016deep, finn2016unsupervised}. We plan with the learned scores and dynamics model with a standard MPC algorithm with random actions that looks ahead 10 steps.

\begin{figure}
\begin{subfigure}{0.25\linewidth}
   \centering
   \includegraphics[width=\linewidth]{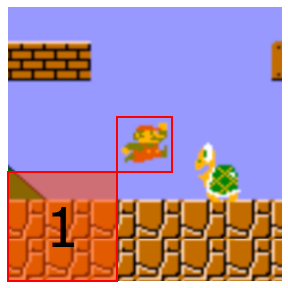}
\end{subfigure}
\begin{subfigure}{0.25\linewidth}
   \centering
   \includegraphics[width=\linewidth]{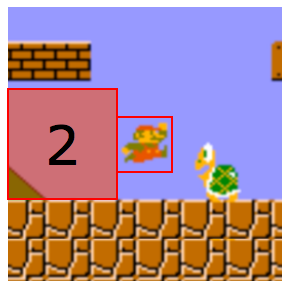}
\end{subfigure}\begin{subfigure}{0.25\linewidth}
   \centering
   \includegraphics[width=\linewidth]{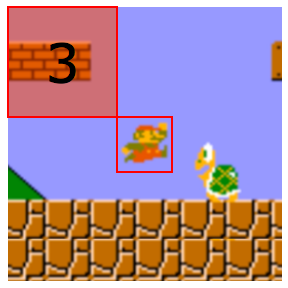}
\end{subfigure}\begin{subfigure}{0.25\linewidth}
   \centering
   \includegraphics[width=\linewidth]{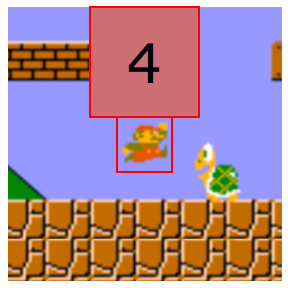}
\end{subfigure}

\begin{subfigure}{0.25\linewidth}
   \centering
   \includegraphics[width=\linewidth]{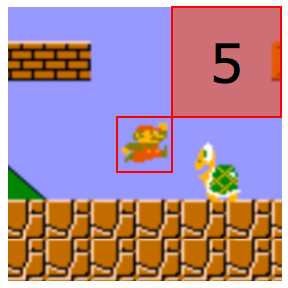}
\end{subfigure}\begin{subfigure}{0.25\linewidth}
   \centering
   \includegraphics[width=\linewidth]{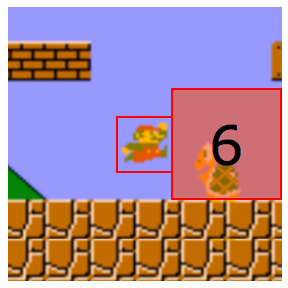}
\end{subfigure}\begin{subfigure}{0.25\linewidth}
   \centering
   \includegraphics[width=\linewidth]{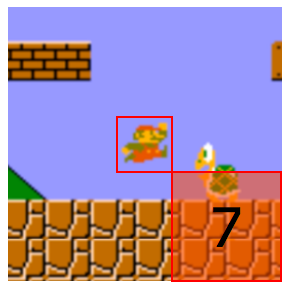}
\end{subfigure}\begin{subfigure}{0.25\linewidth}
   \centering
   \includegraphics[width=\linewidth]{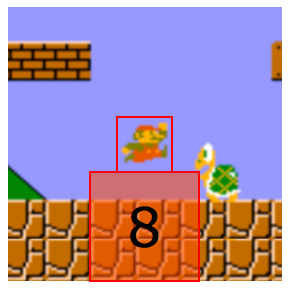}
\end{subfigure}

\begin{subfigure}{0.25\linewidth}
   \centering
   \includegraphics[width=\linewidth]{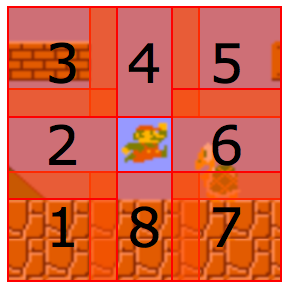}
\end{subfigure}

\caption{A visualization of the sub-regions in the Super Mario Bros game. In this game, there are in total 8 sub-regions.}\label{fig:locations}
\end{figure}

\subsubsection{Architectures for score function and dynamics model.} 
\label{sec:app:mario_arch}
For the score function, we train a CNN taking each 12px by 12px sub-region as input with 2 conv layers and 1 hidden fully connected layers. For each conv layer, we use a filter of size 3 by 3 with stride 2 with number of output channels equals to 8 and 16 respectively. ``Same padding'' is used for each conv layer. The fully connected layers have 128 units.  Relu functions are applied as activation except the last layer. 

For the dynamics model, we train a neural network with the following inputs: a. 30 by 30 egocentric observation around mario. b. current action along with 3 recent actions encoded in one-hot tensor. c. 3 most recent position shifts. d. one-hot encoding of the current planning step. Input a is encoded with 4 sequential conv layers with kernel size 3 and stride 2. Output channels are 8, 16, 32, 64 respectively. A global max pooling follows the conv layers. Input b, c, d are each encoded with a 64 node fc layer. The encoded results are then concatenated and go through a 128 units hidden fc layer. This layer connects to two output heads, one predicting shift in location and one predicting ``done'' with sigmoid activation. Relu function is applied as activation for all intermediate layers.  

\subsubsection{Hyperparameters.} During training, we use an adam optimizer with learning rate 3e-4, $\beta_1=0.9$, $\beta_2 = 0.999$. The batchsize is 256 for score function training and 64 for dynamics model.  We use $horizon=10$ as our planning horizon. We use a discount factor $\gamma=0.95$ and 128 environments in our MPC. We use 1e-4 for the regularization term for matrices entries.

\subsubsection{More on training}
In the scoring function training, each data point is a tuple of a down sampled trajectory and a calculated score. We down sample the trajectory in the exploration data by taking data from every two steps. Half the the  trajectories ends with a ``done''(death) event and half are not. For those ends with ``done'', the score is the distance mario traveled by mario at the end. For the other trajectories, the score is the distance mario traveled by the end plus a mean future score. The mean future score of a trajectory is defined to be the average extra distance traveled by longer (in terms of distance) trajectories than our trajectory. We note that all the information are contained in the exploration data.

\subsubsection{More Details on Baselines}\label{sec:mario_baseline_app}

\textbf{Exploration Data} Exploration Data is the data from which we learn the scores, dynamics model and imitate. The data is collected from a suboptimal policy described in Appendix~\ref{sec:mario_baseline_app}. The average reward on this dataset is a baseline for all other methods. This is omitted in new tasks because we only know the performance in the environment where the data is collected. We train a policy with only curiosity as rewards~\cite{pathak2017curiosity}. However, we early stopped the training after 5e7 steps which is far from the convergence at 1e9 steps. We further added an $\epsilon$-greedy noise when sampling demonstrations with $\epsilon=0.4$ for 20000 episodes and $\epsilon=0.2$ for 10000 episodes.  

\textbf{Behavioral Cloning~\cite{pomerleau1989alvinn, bain1995framework}} Behavioral Cloning (BC) learns a mapping from a state to an action on the exploration data using supervised learning. We use cross-entropy loss for predicting the actions. 

\textbf{Model Based with Human Prior} Model Based with Human Priors method (MBHP) incorporates model predictive control with predefined human priors, which is +1 score if the agent tries to move or jump toward the right and 0 otherwise. MBHP replaces the scoring-aggregation step of our method by a manually defined prior. We note that it might be hard to design human priors in other tasks. As super mario is a deterministic environment, we noticed pure behavior cloning trivially 
get stuck at a tube at the very beginning of level 1-1 and die at an early stage of 2-1. Thus we select action using sampling from output logits instead of taking argmax.

\textbf{DARLA~\cite{higgins2017darla}} DARLA relies on learning a latent state representation that can be transferred from the training environments to the testing environment. It achieves this goal by obtaining a disentangled representation of the environment’s generative factors before learning to act. We use the latent representation as the observations for a behavioral cloning agent. We re-implemented and fine-tuned a disentangled representation of observation space as described in \cite{higgins2017darla} for mario environment. We used 128 latent dimensions in both DAE and beta-VAE with $\beta=0.1$. The VAE is trained with batch size 64, learning rate 1e-4. The DAE is trained with batch size 64 and learning rate 1e-3. We tune until the test visualization on training environment is good. We then train a behavior cloning on the learned disentangled representation to benchmark on both training and testing environment.

\textbf{RUDDER~\cite{arjona2018rudder}} RUDDER proposes to use LSTM to decompose the delayed sparse reward to dense rewards. It first trains a function $f(\tau_{1:T})$ predict the terminal reward of a trajectory $\tau_{1:T}$. It then use $f(\tau_{1:t})-f(\tau_{1:t-1})$ as dense reward at step $t$ to train RL policies. We change policy training to MPC so this reward can be used in zero-shot setting. We re-implemented \cite{arjona2018rudder} on both mario and robot enviroment and fine-tuned it respectively. We down-sample mario trajectory by a factor of 2 to feed into a special LSTM in RUDDER's source code. For mario environment, the feature vector for each time step is derived from a convolution network with channels 1, 3, 8, 16, 32; kernal sizes 3, 3, 3, 3 respectively, followed by a fc layer with 64 output features. The 64 is then feed into each LSTM step. For robot reaching environment,   we directly use a fc layer to down sample observation feature from 3250 to 64 and feed into LSTM. Both models are trained with batch size 256 under learning rate 2e-4, with weight decay of 5e-3.

\textbf{Behavior Clone with Privilege Data} Instead of using exploratory trajectories from the environment, we collect a set of near optimal trajectories in the training environment and train a behavior clone agent from it. Note that this is not a fair comparison with other methods, since this method uses better performing training data. 

\textbf{RL curiosity} We use a PPO~\cite{schulman2017proximal} agent that is trained with curiosity driven reward~\cite{burda2018exploration} and the final sparse reward in the training environment. We limit the training steps to 10M. This is also violating the setting we have as it interacts with the environment. We conduct this experiment to test the generalization ability of a RL agent.

\subsubsection{Additional Ablations}\label{sec:mario_ablation_app}
\textbf{Ablation of Planning Steps}
In this section, we conduct additional ablative experiments to evaluate the effect of the planning horizon in a MPC method.  In Table.~\ref{table:plan}, we see that our method fluctuates a little with different planning steps in a relatively small range and outperforms baselines constantly. In the main paper, we choose $horizon = 10$. We find that when plan steps are larger such as 12, the performance does not improve monotonically. This might be due to the difficult to predict long range future with a learned dynamics model.

% \textbf{Ablation of Aggregating functions} We compare the proposed aggregating function which is a simple summation with other functions such as a modified LSTM~\cite{arjona2018rudder} on the Super Mario Benchmark. In Table~\ref{table:aggregation}, we see that the LSTM-based aggregation cannot learn meaningful scores on the Mario Benchmark. We think the original algorithm is designed for Atari games rather than long-horizon games such as the Mario Bros. 

% \begin{table}
% \centering
% \caption{Ablation Study for number of planning steps in MPC based methods.The averaged return is reported. \label{table:plan}}
% \begin{tabular}{@{}llllclll@{}}\toprule
% & \multicolumn{3}{c}{World 1 Stage 1} & \phantom{ab} & \multicolumn{3}{c}{World 2 Stage 1(new task)}  \\
% \cmidrule{2-4} \cmidrule{6-8}
%           & plan 8  & plan 10 & plan 12 & & plan 8  & plan 10 & plan 12 \\ \midrule
% SAP      & \textbf{1341.5} & \textbf{1359.3} & \textbf{1333.5} & & \textbf{724.2} & \textbf{790.1} & \textbf{682.6}  \\
% MBHP      & 1193.8 & 1041.3 & 1112.1 & & 546.4 &  587.9 & 463.8 \\
% \bottomrule
% \end{tabular}
% \end{table}

\begin{table}
\centering
\caption{Ablation Study for number of planning steps in MPC based methods.The averaged return is reported. \label{table:plan}}
\begin{tabular}{llll}\toprule
& \multicolumn{3}{c}{World 1 Stage 1}\\
\cmidrule{2-4} 
           & plan 8  & plan 10 & plan 12\\ \midrule
SAP      & \textbf{1341.5} & \textbf{1359.3} & \textbf{1333.5}\\
MBHP      & 1193.8 & 1041.3 & 1112.1\\
\bottomrule
\end{tabular}
\begin{tabular}{llll}\toprule
& \multicolumn{3}{c}{World 2 Stage 1(new task)}\\
\cmidrule{2-4} 
           & plan 8  & plan 10 & plan 12\\ \midrule
SAP      & \textbf{724.2} & \textbf{790.1} & \textbf{682.6}\\
MBHP      & 546.4 &  587.9 & 463.8\\
\bottomrule
\end{tabular}
\end{table}

% \begin{table}
% \centering
% \caption{Ablation Study of aggregating function. The averaged return is reported. \label{table:aggregation}}
% \begin{tabular}{@{}lll@{}}\toprule
% & {World 1 Stage 1} & {World 2 Stage 1(new task)}  \\
%             \\ \midrule
% $SAP-subregion$      & \textbf{1359.3} & \textbf{790.1}  \\
% $SAP-RUDDER$     & 1258.0          & 737.0  \\
% \bottomrule
% \end{tabular}
% \end{table}

\subsubsection{Additional visualization}\label{sec:mario_viz_app}
\begin{figure}
\begin{subfigure}{0.32\linewidth}
   \centering
   \includegraphics[width=\linewidth]{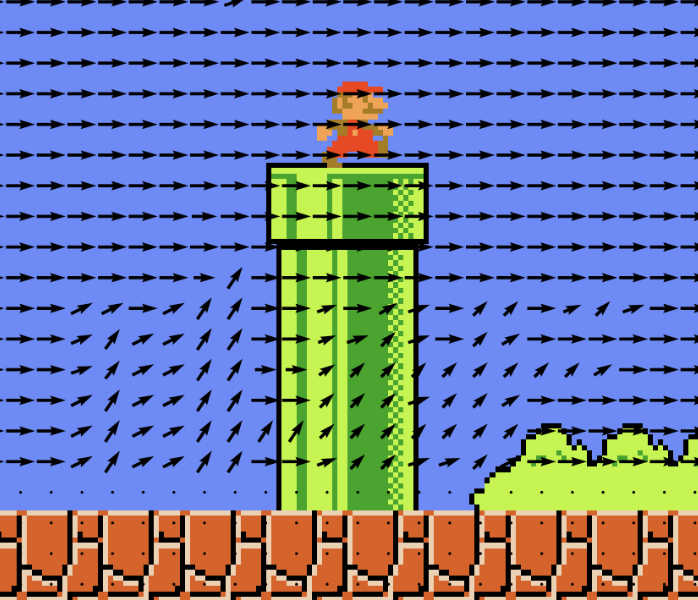}
\end{subfigure}
\begin{subfigure}{0.32\linewidth}
   \centering
   \includegraphics[width=\linewidth]{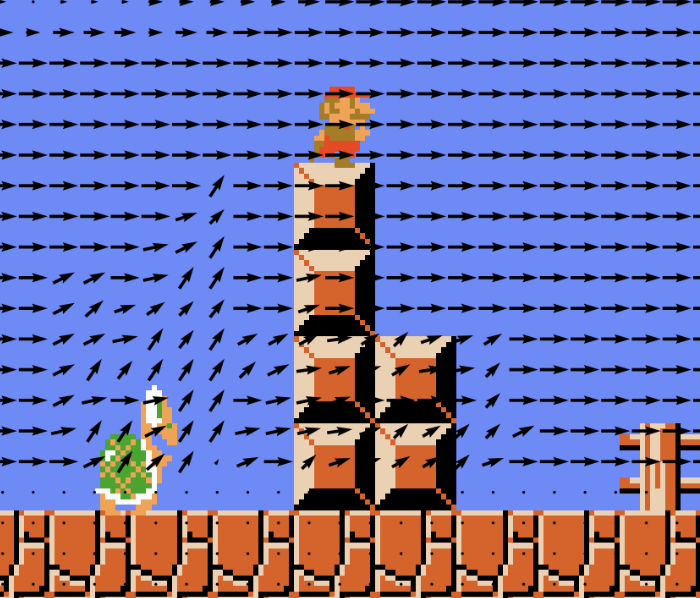}
\end{subfigure}
\begin{subfigure}{0.32\linewidth}
   \centering
   \includegraphics[width=\linewidth]{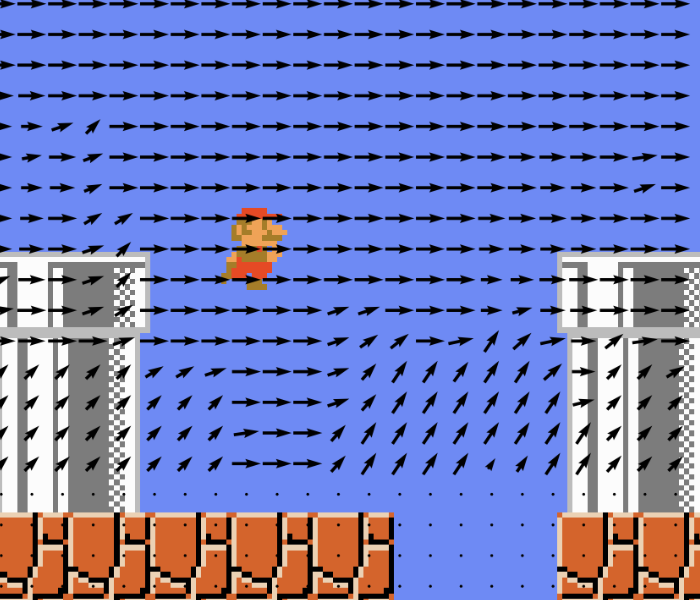}
\end{subfigure}

\caption{More visualizations on the greedy action map on W1S1, W2S1(new task) and W5S1(new task). Note the actions can be different from the policy from MPC.}\label{fig:viz_app}
\end{figure}

In this section, we present additional visualization for qualitative study. In Figure.~\ref{fig:viz_app}, we see that on a few randomly sampled frames, even the greedy action can be meaningful for most of the cases. We see the agent intend to jump over obstacles and avoid dangerous monsters. 

\begin{figure}[h]
\begin{subfigure}{0.24\linewidth}
   \centering
   \includegraphics[width=\linewidth]{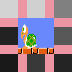}
   \caption{move right}\label{fig:viz_score_a}
\end{subfigure}
\begin{subfigure}{0.24\linewidth}
   \centering
   \includegraphics[width=\linewidth]{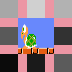}
   \caption{jump}\label{fig:viz_score_b}
\end{subfigure}
\begin{subfigure}{0.24\linewidth}
   \centering
   \includegraphics[width=\linewidth]{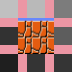}
   \caption{move right}\label{fig:viz_score_c}
\end{subfigure}
\begin{subfigure}{0.24\linewidth}
   \centering
   \includegraphics[width=\linewidth]{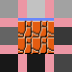}\label{fig:viz_score_d}
   \caption{jump}
\end{subfigure}
\caption{Visualization of the learned score on pre-extracted objects. The grayscale area is the visualized score and the pink area is a separator. Best viewed in color. In (a), we synthetically put the mario agent in 8 relative positions to ``koopa'' conditioned on the action ``move right''. The score is significantly lower when the agent's position is to the left of ``koopa'' compared to other position. In (b), it is the same setup as in (a) but conditioned on the action ``jump''. We find that across (a) and (b) the left position score of (b) is smaller than that of (a) which is consistent with human priors. In (c) and (d), we substitute the object from ``koopa'' to the ground. We find that on both (c) and (d) the score are similar for the top position which means there is not much difference between different actions. Note this figure is only for visualizations and we even put the agent in positions that cannot be achieved in the actual game.}\label{fig:viz_score}
\end{figure}

In Figure.~\ref{fig:viz_score}, we show the scores of a given state-action pair and find that the scores fulfill the human prior. For example, in Figure.~\ref{fig:viz_score}\subref{fig:viz_score_a}, we synthetically put the mario agent in 8 relative positions to ``koopa'' conditioned on the action ``move right''. The score is significantly lower when the agent's position is to the left of ``koopa'' compared to other position. In Figure.~\ref{fig:viz_score}\subref{fig:viz_score_b}, it is the same setup as in Figure.~\ref{fig:viz_score}\subref{fig:viz_score_a} but conditioned on the action ``jump''. We find that across Figure.~\ref{fig:viz_score}\subref{fig:viz_score_a} and Figure.~\ref{fig:viz_score}\subref{fig:viz_score_b} the left position score of Figure.~\ref{fig:viz_score}\subref{fig:viz_score_b} is smaller than that of Figure.~\ref{fig:viz_score}\subref{fig:viz_score_a} which is consistent with human priors. In Figure.~\ref{fig:viz_score}\subref{fig:viz_score_c} and Figure.~\ref{fig:viz_score}\subref{fig:viz_score_d}, we substitute the object from ``koopa'' to the ground. We find that on both Figure.~\ref{fig:viz_score}\subref{fig:viz_score_c} and Figure.~\ref{fig:viz_score}\subref{fig:viz_score_d} the score are similar for the top position which means there is not much difference between different actions. 

\subsubsection{Additional Results with Ground Truth Dynamics Model and No Done Signal}
\begin{figure}[h]
\centering
   \begin{subfigure}{0.48\linewidth}
   \centering
   \includegraphics[width=\linewidth]{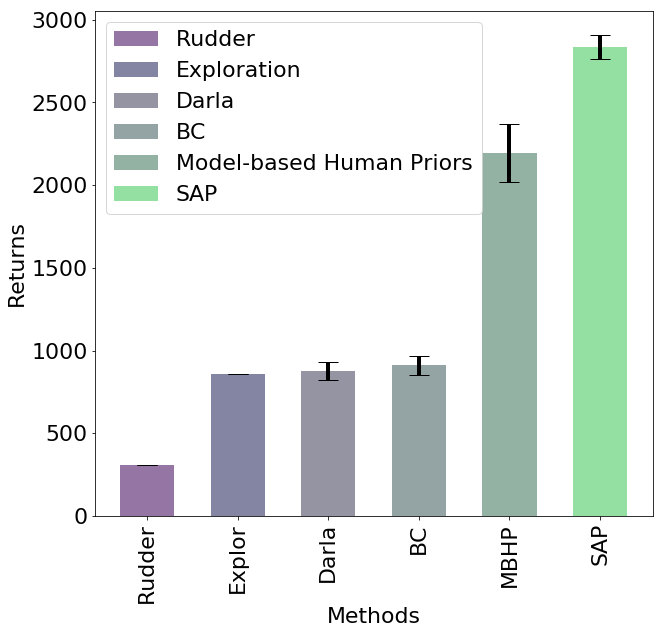}
   \caption{GT dynamics model on W1S1 (train)}
   \label{fig:ablation_a} 
\end{subfigure}
\hfill
\begin{subfigure}{0.48\linewidth}
   \centering
   \includegraphics[width=\linewidth]{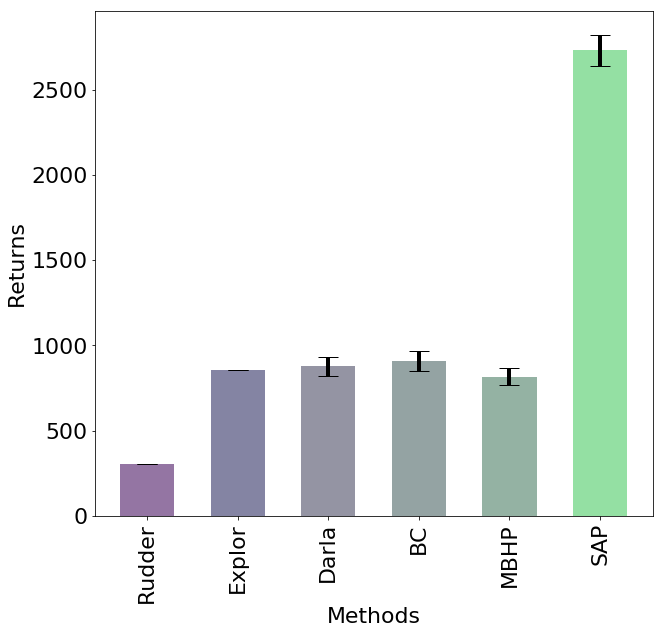}
   \caption{GT dynamics model \& no done on W1S1 (train)}
   \label{fig:ablation_b}
\end{subfigure}
\caption{Training Performance with perfect dynamics and no done signal.}
\end{figure}

\subsection{Robotics Blocked Reach}
\subsubsection{Environment.}\label{sec:robot_env_app}
In the Blocked Reach environment, a 7-DoF robotics arm is manipulated for a specific task. For more details, we refer the readers to \cite{plappert2018multi}. We discretize the robot world into a $200 \times 200 \times 200$ voxel cube. For the action space, we discretize the actions into two choices for each dimension which are moving 0.5 or -0.5. Hence, in total there are 8 actions. We design four configurations for evaluating different methods as shown in Figure~\ref{fig:samples}. For each configurations, there are three objects are placed in the middle as obstacles. The height of the objects in each configuration are (0.05, 0.1, 0.08), (0.1, 0.05, 0.08), (0.12, 1.12, 0.12), (0.07, 0.11, 0.12).

\subsubsection{Applying SAP on the 3D robotic task}
\label{sec:app:sap_robot}
We apply the SAP framework as follows. The original observation is a 25-dimensional continuous state and the action space is a 3-dimensional continuous control. They are discretized into voxels and 8 discrete actions as described in Appendix~\ref{sec:robot_env_app}. In this environment, the egocentric region is set to a $15 \times 15 \times 15$ cube of voxels around robot hand end effector. We divide this cube into 27 $5 \times 5 \times 5$ sub-regions. The scoring function is a fully-connected neural network that takes in a flattened voxel sub-region and outputs the score matrix with a shape of $26 \times 8$. The scores for each step are aggregated by a sum operator along the trajectory. We also train a 3D convolutional neural net as the dynamics model. The dynamics model takes in a $15 \times 15 \times 15$ egocentric region as well as an action, and outputs the next robot hand location. With the learned scores and the dynamics model, we plan using the MPC method with a horizon of 8 steps.  

\subsubsection{Architectures for score function and dynamics model.}
For the score function, we train a 1 hidden layer fully-connected neural networks with 128 units. We use Relu functions as activation except for the last layer. Note that the input 5 by 5 by 5 voxels are flattened before put into the scoring neural network.

For the dynamics model, we train a 3-D convolution neural network that takes in a egocentric region (voxels),  action and last three position changes. The 15 by 15 by 15 egocentric voxels are encoded using three 3d convolution with kernel size 3 and stride 2. Channels of these 3d conv layers are 16, 32, 64, respectively. A 64-unit FC layer is connected to the flattened features after convolution. The action is encoded with one-hot vector connected to a 64-unit FC layer. The last three $\delta$ positions are also encoded with a 64-unit FC layer. The three encoded features are concatenated and go through a 128-unit hidden FC layer and output predicted change in position. All intermediate layers use relu as activation.   

\subsubsection{Hyperparameters.}
During training, we use an adam optimizer with learning rate 3e-4, $\beta_1=0.9$, $\beta_2 = 0.999$. The batchsize is 128 for score function training and 64 for dynamics model.  We use $horizon=8$ as our planning horizon. We use 2e-5 as the weight for matrices entry regularization.

\subsubsection{Baselines}\label{sec:robo_baseline_app}
Our model based human prior baseline in the blocked robot environment is a 8-step MPC where score for each step is the y component of the action vector at that step. 

We omit the Behavioral cloning baselines, which imitates exploration data, as a consequence of two previous results. 

% \subsection{Super Mario Bros}
% \subsection{Blocked Reacher}

%%% ACRONYMS
\begin{acronym}
\acro{HP}{high-pass}
\acro{LP}{low-pass}
\end{acronym}

\end{document}